\documentclass[10pt,twocolumn,letterpaper]{article}

\usepackage[pagenumbers]{cvpr} 

\usepackage[dvipsnames]{xcolor}

\usepackage{times}
\usepackage{epsfig}
\usepackage{graphicx}
\usepackage{amsmath}
\usepackage{booktabs}
\usepackage{amssymb}
\usepackage{pifont}
\usepackage[export]{adjustbox}
\definecolor{cvprblue}{rgb}{0.21,0.49,0.74}
\usepackage[pagebackref,breaklinks,colorlinks,citecolor=cvprblue]{hyperref}

\def\paperID{121} 
\def\confName{3DV\xspace}
\def\confYear{2024\xspace}

\title{Partial-View Object View Synthesis via Filtering Inversion}

\author{Fan-Yun Sun$^{1,2}$\thanks{This work was partly done during an internship at Nvidia.} \and Jonathan Tremblay$^{2}$ \and Valts Blukis$^{2}$
\and
Kevin Lin$^{1}$ \and Danfei Xu$^{2,3}$ \and Boris Ivanovic$^{2}$ \and Peter Karkus$^{2}$ 
\and
Stan Birchfield$^{2}$ \and Dieter Fox$^{2}$ \and  Ruohan Zhang$^{1}$ \and Yunzhu Li$^{1}$
\and
Jiajun Wu$^{1}$ \and Marco Pavone$^{1,2}$ \and Nick Haber$^{1}$ \and \\
$^1$ Stanford University, $^2$ Nvidia,
$^3$ Georgie Institute of Technology\\
{\tt\small fanyun@stanford.edu}
}

\definecolor{darkgreen}{RGB}{0,127,0}
\definecolor{darkred}{RGB}{200,0,0}

\newcommand{\xmark}{\text{\ding{55}}}  
\def\greencheckmark{\textcolor{darkgreen}{\checkmark}}
\def\redxmark{\textcolor{darkred}{\xmark}}

\newcommand{\pk}[1]{}

\def\method{FINV\xspace}

\begin{document}
\maketitle

\vspace{-4mm}
\begin{abstract}
\vspace{-2mm}
We propose Filtering Inversion (\method{}), a learning framework and optimization process that predicts a renderable 3D object representation from one or few partial views. 
\method{} addresses the challenge of synthesizing novel views of objects from \textit{partial} observations, spanning cases where the object is not entirely in view, is partially occluded, or is only observed from similar views.
%
To achieve this, \method learns shape priors by training a 3D generative model.
At inference, given one or more views of a novel real-world object, \method first finds a set of latent codes for the object by inverting the generative model from multiple initial seeds. 
Maintaining the set of latent codes, \method filters and resamples them after receiving each new observation, akin to particle filtering. The generator is then finetuned for each latent code on the available views in order to adapt to novel objects. 
We show that \method successfully synthesizes novel views of real-world objects (e.g., chairs, tables, and cars), even if the generative prior is trained only on synthetic objects. The ability to address the sim-to-real problem allows \method to be used for object categories without real-world datasets.
\method{} achieves state-of-the-art performance on multiple real-world datasets, 
recovers object shape and texture from partial and sparse views, is robust to occlusion, and is able to incrementally improves its representation with more observations.

\end{abstract}

\vspace{-5mm}

\section{Introduction}

\begin{figure}[!tbp]
   \centering
      \includegraphics[width=\linewidth]{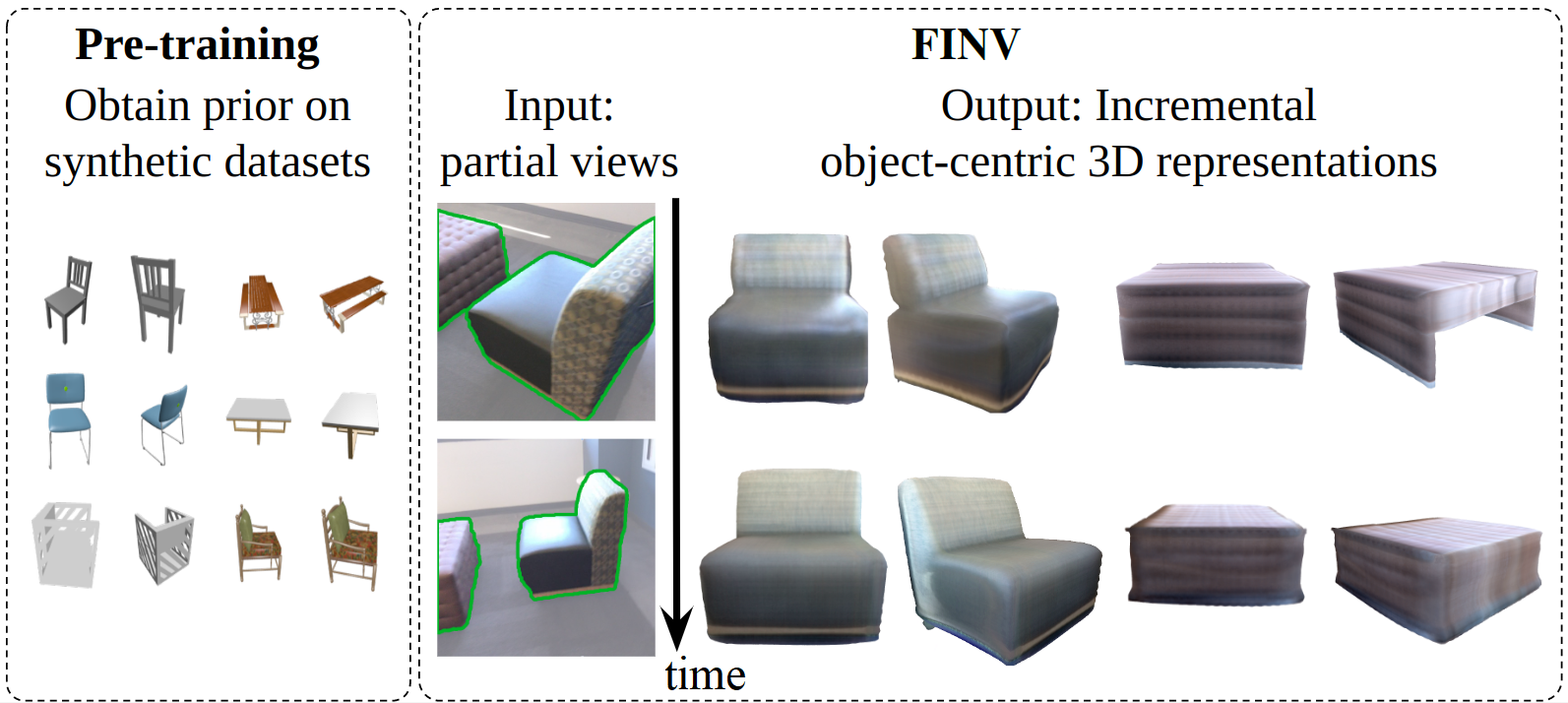}
     \caption{
{\sc Left:}  Our \method method learns a category-level prior by training on synthetic data. 
{\sc Right:} At test time, from one or more posed RGB images with object masks (green contours), our method generates textured meshes. Our model is incremental, yielding better reconstructions as more observations are obtained (top to bottom). 
          Note that the coffee table is only partially observed, and each object is reconstructed independently. 
     }
     \vspace{-4mm}
  \label{fig:example_lf_trajs}
\end{figure}

\label{sec:intro}

\begin{figure}
   \centering
      \includegraphics[width=\linewidth]{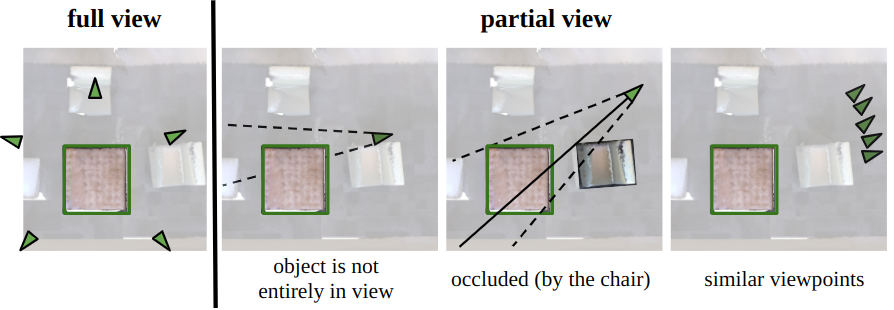}
     \caption{
    Rather than assuming that the object is fully observed (left), 
    we focus on view synthesis under challenging \textit{partial} views, spanning cases where the object is not entirely in view, is partially occluded, or is only observed from only similar viewpoints. 
     }
     \vspace{-4mm}
  \label{fig:partial}
\end{figure}

We study the problem of synthesizing novel views of an object from a sparse set of challenging \textit{partial} views, in which some parts of the object might not be seen by any view (see Figure~\ref{fig:example_lf_trajs}). Sparse-view novel view synthesis has seen recent advancements thanks to neural rendering and learning (object) priors~\cite{yu2021pixelnerf,Kyriazi:objnerf,mueller2022autorf,wang2021ibrnet}. At a high level, these methods produce a neural scene description from few views (1--3) using prior knowledge, then use the scene description to render images from different perspectives. While these methods have shown impressive results, they fail with partial input views where the object is occluded or not fully visible, as it is often the case in practical applications such as robotics.
Additionally, they do not address any domain gap between the training dataset and test-time observations.
To address these issues, we propose Filtering Inversion (\method), a method that learns a category-level object shape and appearance prior from a large variety of instances. Inspired by pivotal tuning~\cite{roich2022pivotal}, at test time we perform a novel two-stage optimization process to first retrieve object latent representations from the generative model and then fine-tune the generator for each latent code. To combat the instability of GAN inversion, \method incorporates a novel filtering and resampling process of latent codes, akin to particle filtering.
As a result, we can handle situations where the object of interest is only partially visible, occluded, or viewed from a limited number of similar perspectives (see Fig.~\ref{fig:partial}).
Unlike previous methods, such as pixelNeRF~\cite{yu2021pixelnerf} and AutoRF~\cite{mueller2022autorf}, 
our method generates a complete 3D mesh that can be used in classical rendering pipelines.\looseness=-1

More specifically, \method uses a 3D GAN model called GET3D~\cite{gao2022get3d} to learn the object prior. We train GET3D on a dataset of objects of the target category and then use it in the real world.
We demonstrate that \method can leverage the ever-growing collection of synthetic data~\cite{chang2015shapenet} for learning object priors.
At test time, our filtering inversion process can close the sim-to-real gap~\cite{tremblay2018training} and generalize to novel real-world object instances.

Our contributions can be summarized as follows:
\begin{itemize}
    \item We propose a framework that combines the strengths of generative modeling~\cite{gao2022get3d} and network fine-tuning~\cite{roich2022pivotal} to generate photorealistic novel view renderings of objects from \textit{partial} RGB views. Our framework can incrementally improve upon the reconstruction from a stream of observations.
    \item We introduce \emph{filtering inversion}, a novel filtering process that facilitates automatic search in the latent space to overcome the instability in the inversion process.
    \item  We show state-of-the-art results on several real-world datasets (including tables, chairs, and cars), demonstrating that our two-phase method is able to synthesize new views of objects in the real world, without having to train on a real-world dataset. 
    We conduct ablation studies showing that our filtering process contribute to the final performance.
\end{itemize}

 \vspace{-2mm}
\section{Related Work}

\vspace{-1mm}
\paragraph{Sparse-View Novel View Synthesis.}
Table~\ref{tab:relatedwork} shows closely related work, focusing on features that are relevant to solving sparse-view novel view synthesis from partial data.
Optimization methods like NeRS~\cite{zhang2021ners} and DS~\cite{goel2022differentiable} do not learn priors and assume a full coverage of the object.
These methods, like pixelNeRF~\cite{yu2021pixelnerf}, are not designed for partial-view scenarios.
AutoRF~\cite{mueller2022autorf} is the most closely related work to ours, but it does not provide the ability to process input data sequentially, does not address the sim-to-real gap, and only tests on one category (cars).
Diffusion-based optimization from an image prior can also be used for object novel view synthesis~\cite{Kyriazi:objnerf} (at the time of this submission, the code was not yet available to conduct a fair comparison.).
It has been observed that since diffusion-based methods use image instead of 3D priors, they sometimes produce 3D inconsistent views, {\em e.g.}, Janus-faced animals with faces on both front and back sides~\cite{metzer2022latent, Kyriazi:objnerf}.
\vspace{-2mm}
\paragraph{Inverse object estimation via generative object priors.}

Recent 3D generative models leverage a hybrid framework that considers both explicit and implicit object representations~\cite{gao2022get3d,chan2022efficient,or2022stylesdf,fan2022unified,zhang20223d,takikawa2022variable}. 
They have shown impressive performance in producing high-quality geometry and detailed texture information.
%
Prior works have also shown that, through test-time optimization, the learned object/scene priors can facilitate inference of the camera pose, shape, and texture~\cite{park2019deepsdf,sitzmann2019scene,jang2021codenerf,yen2021inerf,mueller2022autorf}.
However, these works typically rely on labeled real-world data or have a time-consuming optimization process. The efficacy of many of these methods in real-world scenarios involving \textit{partial} observations is unclear and will be discussed in this paper.
%
%
For more discussion on related works, refer to section~\ref{supp:related} in the supplementary materials.

\begin{table}
\setlength{\tabcolsep}{4pt}
\resizebox{\linewidth}{!}{%
    \begin{tabular}{rcccccc}
    \toprule
    & NeRF & NeRS & DS & pNeRF & AutoRF & FINV \\ \hline
    expected num. images & 100+ & 8--16 & 4--12 & 1--3 & 1 & 1--5 \\
    handles single view         & \redxmark   & \redxmark  & \redxmark & \greencheckmark       & \greencheckmark    & \greencheckmark   \\
    handles occlusion                                & \redxmark   & \redxmark   & \redxmark & \redxmark         & \greencheckmark    & \greencheckmark   \\
    handles not entirely in view & \redxmark   & \redxmark  & \redxmark & \redxmark       & \greencheckmark    & \greencheckmark   \\
     learns (object) prior                                 & \redxmark   & \redxmark   & \redxmark & \greencheckmark       & \greencheckmark    & \greencheckmark   \\
     closes sim-to-real gap                          & --    & --    & -- & \greencheckmark      & \redxmark     & \greencheckmark   \\                  
    produces mesh output$^*$ & \redxmark   & \greencheckmark   & \greencheckmark & \redxmark       & \redxmark    & \greencheckmark \\
    \bottomrule
    \end{tabular}%
}
\caption{\textbf{Comparison with prior work on (sparse) novel view synthesis.} 
NeRF~\cite{mildenhall2020nerf}, NeRS~\cite{zhang2021ners}, DS~\cite{goel2022differentiable}, and pixelNeRF~\cite{yu2021pixelnerf} have difficulty dealing with occlusion, especially when the area of occlusion is unknown. 
Note that pixelNeRF takes spatial image features aligned to each pixel as input, requiring the object to be entirely in view when only one view is given. ($^*$without post processing)
}
\vspace{-3mm}
\label{tab:relatedwork}
\end{table}

 \begin{figure*}[!tbp]
   \centering
      \includegraphics[width=0.7\linewidth]{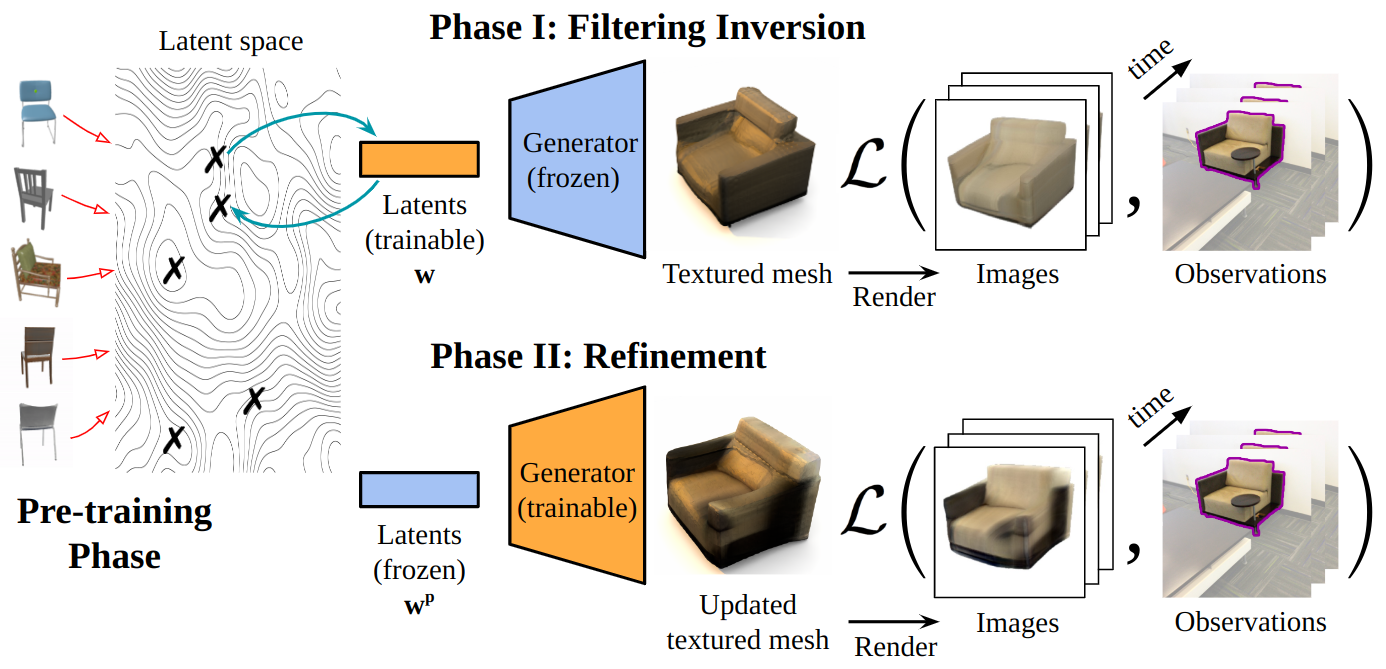}
     \vspace{-1mm}
  \caption{\textbf{Method overview.} Phase I: given a set of observed view(s), our method first samples a set of latent codes and then optimizes those latent codes for creating a 3D model that matches observed view(s), see Figure~\ref{fig:method_phase1}. Phase II: following the latent optimization, we freeze the latent codes and optimize the generator part of the network by fine-tuning on the observations. In each phase, the module highlighted in blue is frozen while the one in yellow is trained.}
  \label{fig:method_all}
\end{figure*}

\begin{figure}[!tbp]
   \centering
      \includegraphics[width=1\linewidth]{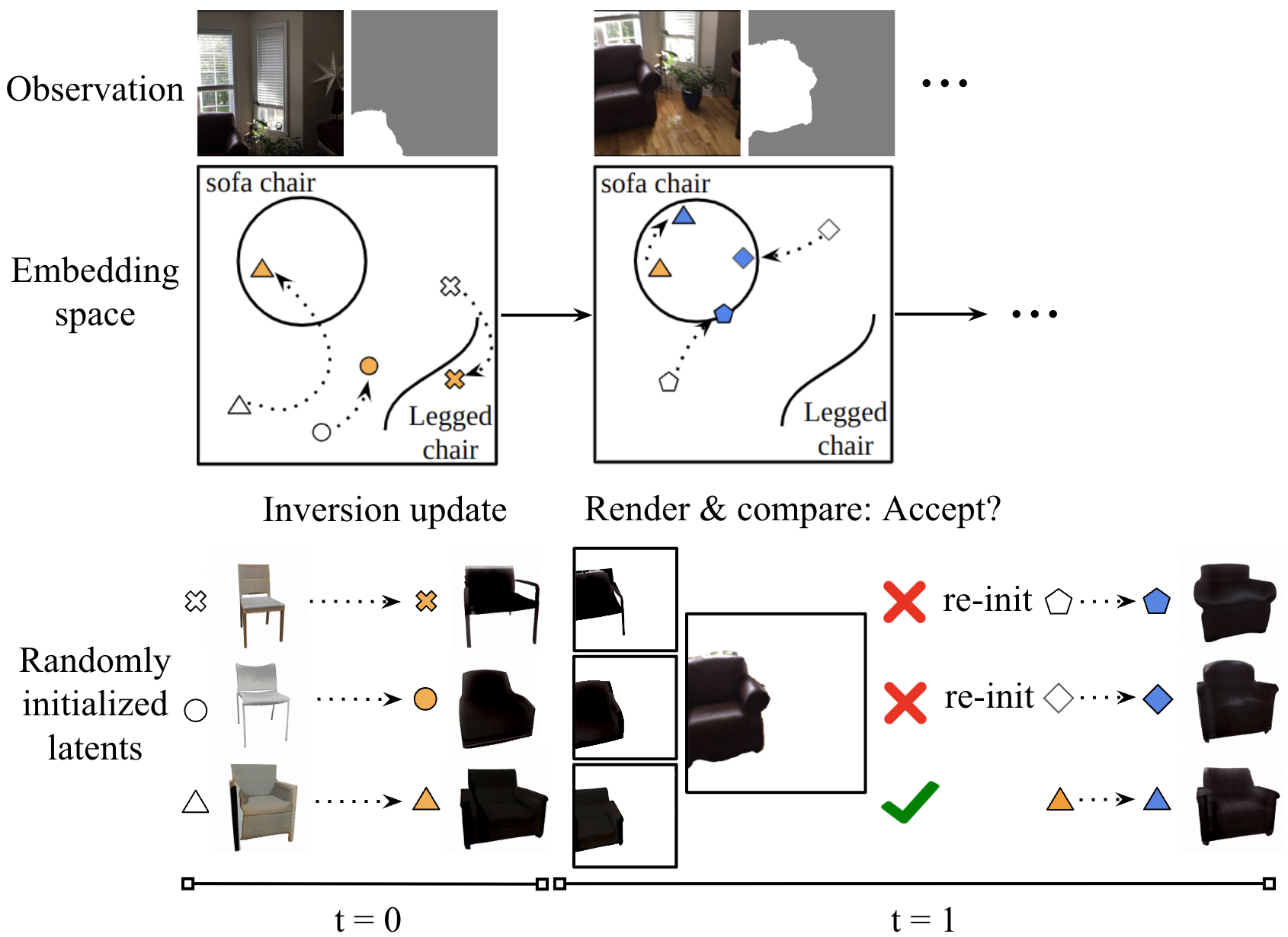}
      \vspace{-1mm}
  \caption{\textbf{Phase I: Filtering Inversion.} The method first samples multiple latent codes (shown by the non-filled icons). 
  Using an inversion update, we refine the sampled latent codes with Eq.~\ref{eq:eq1} (shown by yellow icons).
  Then, we render and compare each latent code using Eq.~\ref{eq:eq2} to decide which ones will be re-sampled or updated further (shown by blue icons).
  }
  \vspace{-2mm}
  \label{fig:method_phase1}
\end{figure}

\section{Method}
In this section, we define the problem and introduce our proposed Filtering Inversion (\method) framework.
The input to our system is a stream of masked RGB observations of an object in a scene, with camera poses.
We denote by $\mathbf{I}_{t}$ and $\mathbf{M}_{t}$ the input image and its corresponding object mask at time $t$, respectively. $\mathbf{I}_{0:T}$ and $\mathbf{M}_{0:T}$ denote the set of images and object masks observed up to time $T$.
In practice, masks and poses can be estimated by off-the-shelf segmentation~\cite{porzi2021improving} and 3D pose estimation and detection models~\cite{ahmadyan2021objectron}. 
More recently methods like~\cite{gkioxari2022cvpr} could be used to obtain scene information needed for our method. 
The task is to, at each time-step $T$, synthesize views of the object that have not been observed up to time $T$ using RGB images $\mathbf{I}_{0:T}$  and object masks and $\mathbf{M}_{0:T}$.
In other words, we evaluate the model's ability to do novel-view synthesis on the object at each time-step.



Our method adapts GAN priors to create a 3D representation of real-world objects. 
For each object category, we train a GAN that can generate 3D representations of objects in the category. 
More specifically, we leverage large-scale synthetic data~\cite{chang2015shapenet} even though that leads to the training and testing data coming from two distinct domains~\cite{tremblay2018training}. 

Given a pre-trained generative object prior, \method uses a two-phase procedure to reconstruct novel objects: a filtering inversion phase, followed by a refinement phase (see Fig.~\ref{fig:method_all}).
The \textbf{filtering inversion} phase serves as a way to iteratively find a latent code that reconstructs the target observations at the given viewpoints.
The \textbf{refinement phase} serves to fit the observations more accurately by fine-tuning the generator model itself while holding the latent code fixed. 
This two-step process is inspired by pivotal tuning inversion~\cite{roich2022pivotal}, which is proposed for latent-space image editing. 
This process 
ensures that, while adapting the model to reconstruct objects in the wild, the distortion of the learned latent space of the 3D GANs is minimized, helping to ameliorate the sim-to-real problem.

In the following, we first introduce the 3D GAN that we use as our backbone generative model. We then introduce \method, our proposed two-phase method that allows us to reconstruct real-world objects efficiently and incrementally.

\subsection{Pretraining Stage}
\method leverages a pre-trained 3D GAN generator $G$; we use GET3D~\cite{gao2022get3d} as the backbone GAN.
%
GET3D is a 3D GAN that disentangles geometry and texture. The geometry branch of the GET3D generator differentiably outputs a surface mesh of arbitrary topology, and the texture branch produces a texture field that can be queried at the surface points to produce texture maps. 
In our ablation studies, we also use EG3D~\cite{Chan2022eg3d} as the backbone and discuss both backbones' advantages and disadvantages. 

At a high level, GET3D samples two input vectors from a Gaussian distribution $\mathbf{z}_\text{geo} \in \mathbb{R}^{512}$ and $\mathbf{z}_\text{tex} \in \mathbb{R}^{512}$. 
Then, following StyleGAN~\cite{karras2021alias,karras2019style,karras2020analyzing}, GET3D uses non-linear mapping networks $f_{\text{geo}}$ and $f_{\text{tex}}$ to map $\mathbf{z}_\text{geo}$ and $\mathbf{z}_\text{tex}$, respectively, to intermediate latent vectors $\textbf{w}_{\text{geo}} = f_{\text{geo}}(\mathbf{z}_\text{geo})$, $\textbf{w}_{\text{tex}} = f_{\text{tex}}(\mathbf{z}_\text{tex})$. 
These intermediate latent vectors are further used to produce a textured mesh.
We use $G_{M}(\textbf{w}_{\text{geo}}; \phi)$ to denote the rendered binary object mask, where $\phi$ represents the parameters of the geometry branch. 
Letting $\textbf{w} = (\textbf{w}_{\text{geo}},\textbf{w}_{\text{tex}})$, we use $G_{I}(\textbf{w}; \phi, \theta)$ to denote the rendered RGB image, where $\theta$ represents the parameters of the texture branch. The mask only depends on the geometry branch, and the final rendering depends on both the geometry and texture branches. 
Note that the image generation also depends on camera intrinsics and extrinsics, which we omit from the above notation for simplicity.




\subsection{Filtering Inversion (Phase I)}
Figure~\ref{fig:method_phase1} shows a high-level workflow of this phase.
Given a pre-trained generator $G$, a set of RGB observations $\mathbf{I}_{0:T}$, and their corresponding object masks $\mathbf{M}_{0:T}$ up to the current time $T$, we aim to optimize a randomly initialized latent code (in practice, we use a set of latent codes) $\textbf{w} = (\textbf{w}_{\text{geo}}, \textbf{w}_{\text{tex}})$ that best encodes the observed object.
In greater detail, we keep the parameters of the generator $G$ fixed and we
optimize with the following objective: 
\begin{align}
\textbf{w}^{p}_{\text{geo}},\textbf{w}^{p}_{\text{tex}}  = 
\underset{\textbf{w} \in \mathbb{R}^{512} \times 2}{\arg\min} \sum_{t} & \left(
\mathcal{L}_{\text{LPIPS}}(\mathbf{I}_{t}, G_{I}(\textbf{w}; \phi, \theta)) \right. +  \nonumber\\
&\left. \mathcal{L}_{\text{MASK}}(\mathbf{M}_{t}, G_{M}(\textbf{w}_{\text{geo}}; \phi) )\right).
\label{eq:eq1}
\end{align}
$\mathcal{L}_{\text{LPIPS}}$ is the LPIPS metric~\cite{zhang2018unreasonable}, and $\mathcal{L}_{\text{MASK}}$ is binary cross entropy in our experiments. We use $\textbf{w}^{p} = (\textbf{w}^{p}_{\text{geo}},\textbf{w}^{p}_{\text{tex}})$ to denote the latent code obtained after the inversion process.

However, the process of inversion can be unstable in practice due to the misalignment between the training and real-world distributions~\cite{song2022editing}.
For example, in our experiments, we train the generator $G$ on synthetic object models, where this problem is particularly pronounced. To combat this issue, we propose filtering inversion, a process that combines inversion with a filtering process akin to particle filtering.

The filtering process is described as follows.  At time $t=0$, we randomly initialize a set of latent codes $\{\textbf{w}_1, ..., \textbf{w}_N\}$ and invert them with Eq.~\eqref{eq:eq1} in parallel, taking image $\mathbf{I}_{0}$ as the reference.
At every following time step $t > 0$, we observe a new image $\mathbf{I}_{t}$ and perform a filtering and optimization step.
In the filtering step, we filter the latent codes, keeping only those that meet the following criterion:
\begin{align}
\text{Rank} \left(\sum_{t} (\mathcal{L}_{\text{LPIPS}}(\mathbf{I}_{t}, G_{I}(\textbf{w}_i; \phi, \theta))\right)  \leq \gamma,
\label{eq:eq2}
\end{align}
where $\text{Rank}(\cdot)$ is the Percentile Ranking that gives a percentage score of the loss out of all the latent codes, and $\gamma$ is a hyper-parameter which we set to $30\%$ in our experiments. In other words, we keep the top $\gamma N$ latent codes that can better reconstruct the object at all views, including the newly observed view $\mathbf{I}_{t}$. 
Next, we proceed to the optimization step, where we again apply Eq.~\eqref{eq:eq1} to update all the latent codes on all the observations up to and including the new observation $\mathbf{I}_{t}$.

We then proceed to time $t+1$ and repeat the filtering and optimization steps.
This filtering and optimization process is executed repeatedly to discard latent codes stuck in local minima and keep improving our representation of the object with every new observation.

\begin{figure*}[!tbp]
   \centering
      \includegraphics[width=\linewidth]{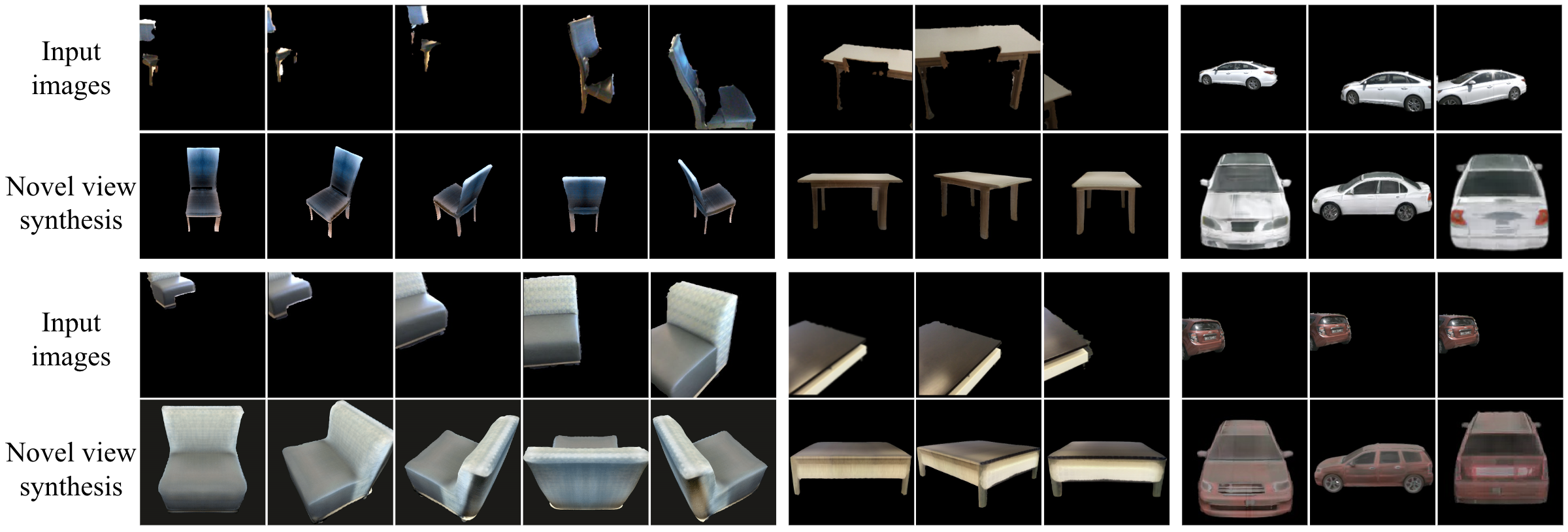}
     \caption{Visualizations of our FINV rendering novel views of various objects given multiple (3 or 5) source views.}
    \vspace*{-3mm}
  \label{fig:vis}
\end{figure*}

\subsection{Refinement (Phase II)}
This phase aims to refine the representations by fine-tuning the generator to reconstruct objects while maintaining the essential priors learned by the generative model. Since our goal is to reconstruct objects in the wild, we will inevitably run into instances with geometry or texture that have not been observed before in the training set. In such cases, simply having the filtered inversion phase (phase I) will be insufficient, as there may not exist a latent code in the learned latent space that enables the generator to produce the required texture and geometry.
If we simply use a random or mean latent code and fine-tune the entire generator around it, the representation tends to overfit to reconstructing the observed views and produces highly-distorted novel views~\cite{roich2022pivotal}. Instead, we fine-tune the generator while fixing $\textbf{w}^{p}$, which can be used to generate a textured mesh that is similar to the observed object in real life. In this way, we expect to be able to minimize the ``distortion'' of the originally well-behaved latent space.

One key insight here is that geometry and texture are of different complexities for different object categories. 
The disentangled architecture of GET3D allows us to fine-tune the geometry and texture of the object separately and with different amounts. 
For example, cars have fewer geometric variations than texture variations, while wooden tables come in many different shapes but have more uniform textures. Thus, we can impose more regularization on the training of geometry to avoid overfitting to the imperfect and partial observations and distorting the learned latent space.

Let $\textbf{w}^{p} = (\textbf{w}^{p}_{\text{geo}}, \textbf{w}^{p}_{\text{tex}})$ be a latent code obtained after phase I inversion at time $t$. We fine-tune the generator using the following objectives:
\begin{align}
\phi^{p} = \underset{\phi}{\arg\min} \sum_{t} &\mathcal{L}_{\text{MASK}}(\mathbf{M}_{t}, G_{M}(\textbf{w}^{p}_{\text{geo}}; \phi)),\label{eq:eq2-1}\\
\theta^{p} = \underset{\theta}{\arg\min} \sum_{t} &\left( \,\mathcal{L}_{\text{LPIPS}}(\mathbf{I}_{t}, G_{I}(\textbf{w}^p; \phi, \theta)) + \right.  \nonumber\notag\\
& \: \lambda_{\text{MSE}} \mathcal{L}_{\text{MSE}}(\mathbf{I}_{t}, G_{I}(\textbf{w}^p; \phi, \theta)) \left.\right)\label{eq:eq2-2},
\end{align}
where $\mathcal{L}_{\text{MSE}}$ is the MSE calculated on pixels and $\lambda_{\text{MSE}}$ is a coefficient.
Recall that $\phi$ and $\theta$ represent the parameters for the generator's geometry and texture branch, respectively. By having the geometry and texture branch's objectives disentangled, we can optimize $\phi$ and $\theta$ to different degrees. In our experiments, we observe that reconstructing geometry tends to be easier than texture and we simply use early-stopping as a regularization on Eq.~\eqref{eq:eq2-1} during the refinement phase.

\section{Experiments}
We conducted extensive experiments on novel view synthesis to evaluate our method. We first compare our method with recent competitive baselines and then conduct an ablation study to understand the effectiveness of our design choices. Additionally, we evaluate our model on 3D shape reconstruction. Finally, we conduct a deeper analysis to understand the relation between performance and the amount of rotation between the source and target view.

\subsection{Setup}
All methods are evaluated on real-world images. We assume a sequential observation of RGB images and instance segmentation masks from a camera moving in the scene relative to the object.
We evaluate the model with either the first 1, 3, or 5 frames as input and report reconstruction quality metrics on frames 6 and onward.
We report reconstruction quality metrics given the first 1, 3, and 5 frames of the object. This allows us to assess the ability of each method to incorporate information from multiple frames in an online setting.
Note that we do not use depth information but assume known camera poses and intrinsic. In this paper, methods compared within the same table are provided with the same estimated poses and segmentations.

\vspace{-2mm}
\paragraph{Metrics.} For novel view synthesis, we report PSNR, SSIM~\cite{wang2004image}, and LPIPS~\cite{zhang2018unreasonable} reconstruction quality metrics.
PSNR measures reconstruction quality at the pixel level, while SSIM and LPIPS take into account semantic perceptual similarity. For the additional experiment on shape reconstruction, we use Chamfer distance and F1 score, following \cite{gkioxari2019mesh}. 

\vspace{-2mm}
\paragraph{Datasets.}
We evaluate all methods on three categories of objects: \textit{Chair}~\cite{dai2017scannet}, \textit{Table}~\cite{dai2017scannet}, and \textit{Car}~\cite{caesar2020nuscenes}. For all categories, we first pre-train each method on ShapeNet~\cite{chang2015shapenet}, a dataset of synthetic objects, to obtain a category-level prior. For \textit{Chair} and \textit{Table}, we evaluate models on ScanNet~\cite{dai2017scannet}, a dataset of real-world indoor scene scans. For \textit{Car}, we evaluate models on NuScenes~\cite{caesar2020nuscenes}, a driving dataset with 3D detection and tracking annotations. For more details on these datasets and the reasons behind their selection, please refer to Section~\ref{supp:dataset} of the supplementary material.

\begin{table*}
\centering
\resizebox{\linewidth}{!}{
\setlength{\tabcolsep}{5pt}
\begin{tabular}{@{}l c c c  c c c  c c c   c c c  c c c  c c c }
\toprule
 & \multicolumn{9}{c}{\textbf{ScanNet Chairs}} & \multicolumn{9}{c}{\textbf{ScanNet Tables}} \\
\cmidrule(lr){2-10}\cmidrule(lr){11-19}
\multicolumn{1}{c}{} & \multicolumn{3}{c}{PSNR$\uparrow$} & \multicolumn{3}{c}{SSIM$\uparrow$} & \multicolumn{3}{c}{LPIPS$\downarrow$}  & \multicolumn{3}{c}{PSNR$\uparrow$} & \multicolumn{3}{c}{SSIM$\uparrow$} & \multicolumn{3}{c}{LPIPS$\downarrow$}   \\
 \cmidrule(lr){2-4}\cmidrule(lr){5-7}\cmidrule(lr){8-10}\cmidrule(lr){11-13}\cmidrule(lr){14-16}\cmidrule(lr){17-19}
 \textbf{\# views} & 1 & 3 & 5 & 1 & 3 & 5 & 1 & 3 & 5 & 1 & 3 & 5& 1 & 3 & 5& 1 & 3 & 5\\
\midrule
Instant-NGP & 21.36 & 22.64 & 23.23 & 0.816 & 0.853 & 0.888 & 0.261 & 0.211 & 0.177 & 15.78 & 16.63 & 17.05 & 0.583 & 0.744 & 0.770 & 0.469 & 0.394 & 0.337\\
pixelNeRF                & 21.47 & 21.58 & 21.96 & 0.687 & 0.774 & 0.847 & 0.284 & 0.234 & 0.187 & 14.83 & 16.59 & 16.96 & 0.592 & 0.699 & 0.731 & 0.415 & 0.325 & 0.306\\
IBRNet     & 22.96 & 23.45 & 23.95 & 0.790 & 0.817 & 0.848 & 0.215 & 0.201 & 0.178 & 15.11 & 16.13 & 16.58 & 0.797 & 0.809 & 0.814 & 0.283 & 0.267 & 0.263 \\
IBRNet + test time opt.  & 22.96 & 23.32 & 23.47 & 0.790 & 0.835 & 0.853 & 0.215 & 0.193 & 0.186 & 15.11 & 16.14 & 16.37 & 0.797 & 0.815 & 0.825 & 0.283 & 0.268 & 0.262\\ 
EG3D + PTI & 20.89 & 22.97 & 24.49 & 0.738 & 0.843 & 0.858 & 0.199 & 0.123 & 0.098 & 17.92 & 19.44 & 19.86 & 0.769 & 0.855 & 0.852  & 0.240 & 0.156 & 0.149  \\
GET3D + PTI & 23.24 & 23.62 & 24.29 & 0.872 & 0.874 & 0.874 & 0.116 & 0.111 & 0.106 & 19.21 & 19.51 & 19.93 & \textbf{0.909} & 0.921 & 0.924 & \textbf{0.163} & 0.138 & 0.130  \\
AutoRF & 22.44 & 22.80 & 22.94 & 0.798 & 0.812 & 0.817 & 0.220 & 0.210 & 0.209 & 13.90 & 14.23 & 14.39 & 0.525 & 0.541 & 0.556 & 0.495 & 0.476 & 0.470\\
FINV-GET3D (Ours) & \textbf{24.61} & \textbf{24.96} & \textbf{26.23} & \textbf{0.937} & \textbf{0.944} & \textbf{0.950}  & \textbf{0.102} & \textbf{0.089} & \textbf{0.082} & \textbf{19.26} & \textbf{19.84} & \textbf{20.48} & 0.907 & \textbf{0.925} & \textbf{0.930} & \textbf{0.163} & \textbf{0.131} & \textbf{0.120} \\
\bottomrule
\end{tabular}
}

\vspace{0.5em}

\caption{\textbf{Results on ScanNet Chairs and Tables.} View synthesis quality for various methods when given 1, 3, or 5 source views from ScanNet chairs and tables. Our \method{} outperforms other methods on a variety of metrics.
}
\label{table:chair-table-results}
\end{table*}

\begin{figure}[!tbp]
   \centering
      \includegraphics[width=1\linewidth,valign=t]{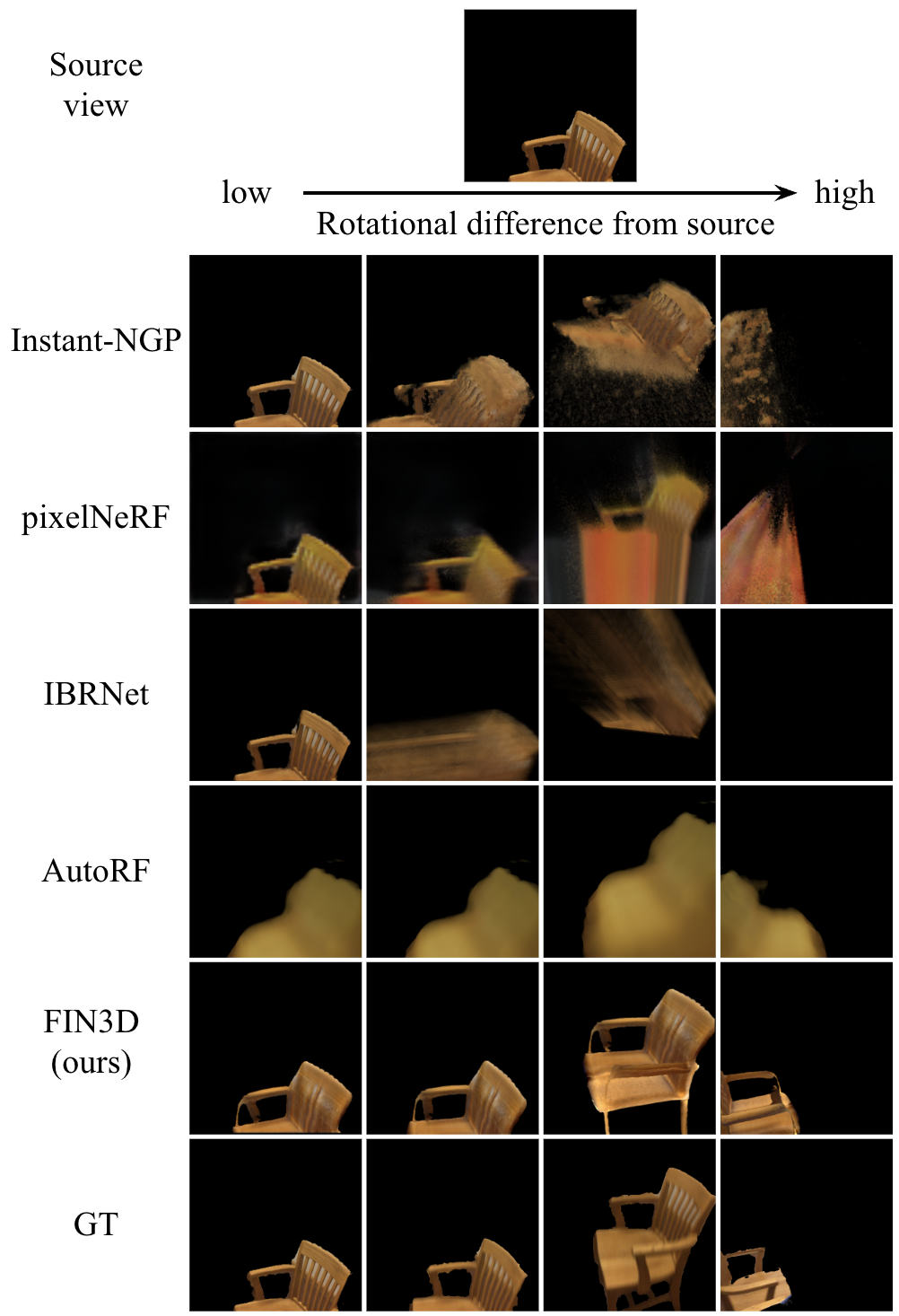}
     \caption{Visualization of our method (FINV) and baselines on a ShapeNet chair, viewed from different perspectives, after training on a single image.} 
    \label{fig:vis-ours-baselines}
   \vspace{-3mm}
\end{figure}

\vspace{-1mm}
\paragraph{Baselines.}
For partial-view novel view synthesis, we compare \method{} against 
Instant-NGP~\cite{muller2022instant},
pixelNeRF~\cite{yu2021pixelnerf}, 
IBRNet~\cite{wang2021ibrnet},
IBRNet fine-tuned during test time, AutoRF~\cite{mueller2022autorf}, and
3D GANs with Pivotal Tuning Inversion (EG3D+PTI and GET3D+PTI)~\cite{Chan2022eg3d,gao2022get3d,roich2022pivotal}. We use open-source implementations of each method when available, or private implementations shared by the authors. 
Note that AutoRF is originally proposed for reconstruction from one single view, but it supports multiple input views as well, which we find to improve reconstruction quality.
Of these, Instant-NGP requires no pretraining; PixelNeRF, IBRNet, AutoRF, the EG3D models used in the baseline EG3D+PTI, and the GET3D models used in our method are pretrained on the synthetic dataset mentioned above. $\text{AutoRF}^{*}$ uses the same architecture as $\text{AutoRF}$, but is pretrained on the NuScenes Car dataset with in-domain data in the same way as the original paper~\cite{mueller2022autorf}.

For shape reconstruction experiments, we compare our method against two state-of-the-art sparse-view reconstruction methods (i.e., NeuS~\cite{wang2021neus} and NeRS~\cite{zhang2021ners}), additional to 3D GANs with Pivotal Tuning Inversion.
Most systems in the above novel view synthesis experiments are either inadequate for the purpose of obtaining shapes/meshes or lack an implementation for extracting them.

\vspace{-2mm}
\paragraph{Implementation Details.}
We run our method on a Linux machine with NVIDIA A40 GPUs. In our experiments, we run 350 gradient update steps in the filtering inversion phase (Phase I).
In the refinement phase, we run 500 gradient update steps.
We ensure that our test-time optimized baselines (Instant-NGP and IBRNet) are optimized with at least the same compute time used for our model. Refer to section~\ref{supp:runtime} for runtime analysis of our method.

\subsection{Results}
Table~\ref{table:chair-table-results} shows the results on ScanNet~\cite{dai2017scannet} chairs and tables, showing the PSNR, SSIM~\cite{wang2004image}, and LPIPS~\cite{zhang2018unreasonable} metrics for 1, 3 and 5 input views.
Our method outperforms all prior work across all metrics for all numbers of input views.
For example, on ScanNet \textit{Chair}s with five input views, we outperform the most competitive prior work in each case by 7.1\%, 7.0\%, and 16.3\% relative improvement in PSNR, SSIM, and LPIPS, respectively.
In the most challenging single-view setting, this improvement increases to 7.2\%, 17.4\%, and 48.7\%, respectively, demonstrating the superior ability of our method to effectively use the learned prior to reconstruct objects from partial views.
The results on ScanNet \textit{Table} data repeat this finding.

Table~\ref{table:car-results} shows the results on NuScenes cars. 
As before, our method consistently outperforms all other methods that were pre-trained on synthetic ShapeNet data across all numbers of input views.
Specifically, we show 2.9\%, 6.4\%, and 8.3\% relative improvement in PSNR, SSIM, and LPIPS, respectively, compared to the most competitive baseline in each case in the single-view setting.
With 5 views, we still outperform all baselines in SSIM and LPIPS; however, Instant-NGP reports higher PSNR.
Since LPIPS is the metric that more accurately reflects human perception~\cite{zhang2018unreasonable,yu2021pixelnerf}, this points to the fact that despite not reconstructing cars the best on a pixel level, our method still gives semantically better reconstructions on target views.
We explore this finding deeper in Section~\ref{sec:ablations}, showing that our model actually achieves a better overall reconstruction.
Surprisingly, despite using only synthetic training data, we achieve comparable results to $\text{AutoRF}^{*}$ (bottom row), which was trained on in-domain real-world images taken from the same distribution as the testing images, in all metrics except multi-view LPIPS.
This demonstrates the effectiveness of our method in bridging the substantial sim-to-real gap. \looseness=-1

\begin{table}
\centering
\resizebox{\linewidth}{!}{
\setlength{\tabcolsep}{3pt}
\begin{tabular}{@{}l  c c c  c c c  c c c }
\toprule
 & \multicolumn{3}{c}{PSNR$\uparrow$} & \multicolumn{3}{c}{SSIM$\uparrow$} & \multicolumn{3}{c}{LPIPS$\downarrow$}  \\
 \cmidrule(lr){2-4}\cmidrule(lr){5-7}\cmidrule(lr){8-10}
 \textbf{\# views} & 1 & 3 & 5 & 1 & 3 & 5 & 1 & 3 & 5 \\
\midrule
\quad Instant-NGP & 14.56 & 15.57 & \textbf{17.19} & 0.580 & 0.623 & 0.647 & 0.546 & 0.519 & 0.489                 \\
\midrule
\multicolumn{3}{l}{pre-trained on synthetic data}\\
\quad pixelNeRF & 12.23 & 13.06 & 13.27 & 0.481 & 0.515 & 0.519 & 0.786 & 0.762 & 0.649\\
\quad IBRNet   & 12.12 & 13.41 & 14.81 & 0.537 & 0.577 & 0.599 & 0.663 & 0.632 & 0.614  \\
\quad IBRNet + test time opt. & 12.12 & 12.89 & 15.60 & 0.537 & 0.517 & 0.636 & 0.663 & 0.646 & 0.566\\ 
\quad EG3D + PTI & 14.33 & 15.43 & 16.50 & 0.606 & 0.647 & 0.670 & 0.564 & 0.517 & 0.487 \\
\quad GET3D + PTI & 13.77 & 15.42 & 16.03 & 0.602 & 0.662 & 0.682 & 0.549 & 0.485 & 0.457 \\
\quad AutoRF & 11.81 & 12.00 & 12.10 & 0.546 & 0.550 & 0.552 & 0.778 & 0.775 & 0.775\\
\quad FINV-EG3D (Ours) & 14.57 & \textbf{15.69} & 16.73 & 0.622 & 0.657 & 0.678 & 0.550 & 0.507 & 0.477 \\
\quad FINV-GET3D (Ours)  &\textbf{14.75} & 15.66 & 16.56 & \textbf{0.645} & \textbf{0.676} & \textbf{0.699} & \textbf{0.517} & \textbf{0.472} & \textbf{0.438} \\
\midrule
\multicolumn{3}{l}{pre-trained on real-world cars}\\
\quad $\text{AutoRF}^{*}$ & 15.11 & 15.76 & 16.32 & 0.663 & 0.677 & 0.698  & 0.656 & 0.642 & 0.698     \\
\bottomrule
\end{tabular}
}
\caption{\textbf{Results on NuScenes Cars.} View synthesis quality for various methods when given 1, 3, 5 source views from NuScenes Cars. Our \method{} outperforms other methods on a variety of metrics.
}
\vspace{-2mm}

\label{table:car-results}
\end{table}
\begin{table}
\centering
\resizebox{.8\linewidth}{!}{
\setlength{\tabcolsep}{5pt}
\begin{tabular}{@{}l   cc cc cc }
\toprule
\textbf{Chairs+Tables} & \multicolumn{3}{c}{Chamfer L2$\downarrow$} & \multicolumn{3}{c}{F1 Score$\uparrow$} \\
 \cmidrule(lr){2-4}\cmidrule(lr){5-7}
 \textbf{\# views} & 1 & 3 & 5 & 1 & 3 & 5 \\
\midrule
NeuS &  0.138 & 0.091 & 0.100 & 0.323 & 0.290 & 0.376 \\
NeRS &  0.157 & 0.076 & 0.095 & 0.438 & 0.452 & 0.448 \\
EG3D + PTI & 0.068 & 0.130 & 0.015 & 0.241 & 0.237 & 0.278 \\
GET3D + PTI & 0.036 & 0.010 & 0.015 & 0.346 & 0.287 & 0.309 \\
\midrule
\method-EG3D  &  0.065 & 0.122 & 0.132 & 0.456 & 0.654 & 0.666 \\
\method-GET3D &  \textbf{0.023} & \textbf{0.009} & \textbf{0.009} & \textbf{0.543} & \textbf{0.676} & \textbf{0.675}  \\
\bottomrule
\end{tabular}
}
\caption{\textbf{Shape reconstruction results on ScanNet.} Shape reconstruction quality for various methods when given 1, 3, or 5 source
views from ScanNet chairs and tables.}
\vspace{-6mm}
\label{table:shape_reconstruction}
\end{table}

\begin{figure}[!tbp]
   \centering
      \includegraphics[width=1\linewidth,valign=t]{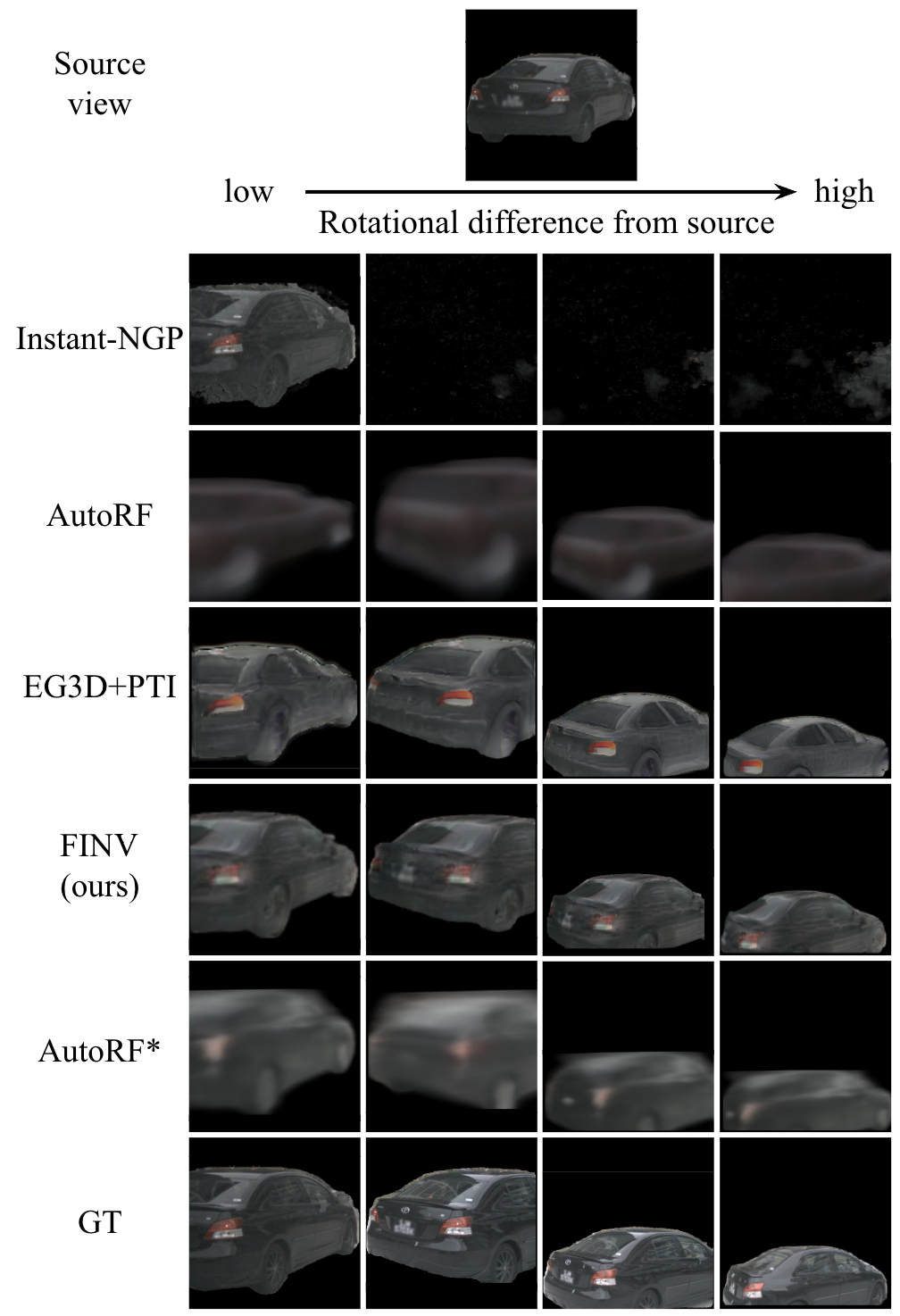}
            \vspace{1mm}
     \caption{Visualization of our method (FINV) and baselines on a NuScenes car, after training on a single view.  Our method preserves better detail, such as the license plate.} 
     \vspace{-6mm}
  \label{fig:vis-ours-baselines2}
\end{figure}

Figures~\ref{fig:vis-ours-baselines} and \ref{fig:vis-ours-baselines2} present qualitative results, highlighting the challenge of our setting: objects are often partially observed and not in full view. 
Although the baselines are able to generate reasonable reconstructions when the target viewpoint is close to the input view, they fail to render unseen parts of the object from viewpoints far from the input.
%
%
We hypothesize that this is because pixelNeRF and IBRNet learn local priors since they are optimized to render from local pixel features. Although IBRNet uses a ray transformer so that density predictions on individual \textit{rays} are coherent, it still learns local priors since the rays are organized in patches.
In contrast, 
we limit our method to update within the latent space of a learned generative model. The learned structure of the latent space serves as regularizer and imposes a ``global prior".
Intuitively, during the filtered inversion phase, we are asking the model to answer the question: ``given these observations (constraints), what would the entire object look like?". 
Subsequently, in the refinement phase, we ask the model to better adapt its answer to the above question to real-world observations.
This two-stage process results in more globally coherent reconstructions.
AutoRF also imposes a global prior, but it uses an encoder network instead of filtered inversion to compute the latent codes. The encoder overfits to its training distribution in the source domain, yielding simulated-looking renderings.
Successful results with AutoRF are only achieved when trained on target-domain data ($\text{AutoRF}^{*}$).
Figure~\ref{fig:vis} shows additional qualitative results.

\vspace{-2mm}
\paragraph{Shape Reconstruction Results.}
We additionally evaluate \method's shape reconstruction quality by calculating the Chamfer L2 Distance and F1 Score between the output mesh with the ground truth object point clouds in ScanNet. From Table~\ref{table:shape_reconstruction}, we observe that \method outperforms methods such as NeuS and NeRS due to the effective use of object priors. Due to the use of Filtered Inversion, \method produces more precise surfaces when compared with EG3D+PTI which leverages similar object priors. \looseness=-1


\vspace{-1mm}
\subsection{Ablations and Analyses}
\label{sec:ablations}
\vspace{-1mm}
In addition to the main experiments shown above, we perform additional experiments to answer the following questions:
\textbf{Q1}: Do FINV's filtering process in the filtered inversion phase and the use of the refinement phase boost performance?
\textbf{Q2}: How does the choice of a mesh-based backbone GAN (\eg, GET3D) compare with radiance field-based backbone GANs (\eg, EG3D)? 
\textbf{Q3}: How does FINV handle rotational deltas between input and target views compared to baselines? 

\begin{table}
\centering
\resizebox{\linewidth}{!}{
\setlength{\tabcolsep}{5.5pt}
\begin{tabular}{@{}l   cc cc cc }
\toprule
 & \multicolumn{6}{c}{PSNR$\uparrow$} \\
\cmidrule{2-7}
 \textbf{\# views} & \multicolumn{2}{c}{1} & \multicolumn{2}{c}{3 }& \multicolumn{2}{c}{5} \\
\midrule
EG3D inv.   &   \multicolumn{2}{c}{17.15 $\pm$ 0.56} & \multicolumn{2}{c}{17.37 $\pm$ 0.48} & \multicolumn{2}{c}{18.13 $\pm$ 0.35}  \\
EG3D inv. w/ filt.   & \multicolumn{2}{c}{17.15 $\pm$ 0.56} & \multicolumn{2}{c}{20.57 $\pm$ 0.32} & \multicolumn{2}{c}{21.19 $\pm$ 0.30}    \\
EG3D (refine. only) & \multicolumn{2}{c}{18.12 $\pm$ 0.52} &  \multicolumn{2}{c}{18.80 $\pm$ 0.34} &  \multicolumn{2}{c}{19.23 $\pm$ 0.37}  \\
EG3D inv. w/ filt. + refine. & \multicolumn{2}{c}{20.35 $\pm$ 0.53} & \multicolumn{2}{c}{22.05 $\pm$ 0.14} & \multicolumn{2}{c}{23.26 $\pm$ 0.18}\\
\midrule 

GET3D inv. &    \multicolumn{2}{c}{21.87 $\pm$ 0.36}  & \multicolumn{2}{c}{22.33 $\pm$ 0.20} & \multicolumn{2}{c}{22.66 $\pm$ 0.16} \\
GET3D inv. w/ filt. & \multicolumn{2}{c}{21.87 $\pm$ 0.36} & \multicolumn{2}{c}{\textbf{23.18 $\pm$ 0.09}} & \multicolumn{2}{c}{23.68 $\pm$ 0.04} \\
GET3D (refine. only) & \multicolumn{2}{c}{19.40 $\pm$ 0.50}  & \multicolumn{2}{c}{20.07 $\pm$ 0.44} & \multicolumn{2}{c}{20.43 $\pm$ 0.49} \\
GET3D inv. w/ filt. +  refine.  &  \multicolumn{2}{c}{\textbf{22.71 $\pm$ 0.27}} & \multicolumn{2}{c}{\textbf{23.19 $\pm$ 0.08}} &   \multicolumn{2}{c}{\textbf{24.07 $\pm$ 0.06}} \\
\midrule
& \multicolumn{3}{c}{SSIM$\uparrow$} & \multicolumn{3}{c}{LPIPS$\downarrow$} \\
\cmidrule(lr){2-4}\cmidrule(lr){5-7}
 \textbf{\# views} & 1 & 3 & 5 & 1 & 3 & 5 \\
\midrule
EG3D inv.    &  0.841 &  0.862 &  0.890 &  0.251 & 0.233 & 0.222 \\
EG3D inv. w/ filt. &  0.841 & 0.868 & 0.902 & 0.251 & 0.182 & 0.169 \\
EG3D (refine. only) & 0.889 & 0.900 & 0.905 & 0.187 & 0.169 & 0.161\\
EG3D inv. w/ filt. + refine.  & 0.793 & 0.889 & 0.885 & 0.193 & 0.119 & 0.104 \\
\midrule
GET3D inv.     & 0.922 & 0.932 & 0.937 & 0.141 & 0.130 & 0.124  \\
GET3D inv. w/ filt.  & 0.922 & \textbf{0.941} &\textbf{0.944} & 0.141 & 0.115 & 0.108  \\
GET3D (refine. only) & 0.913 & 0.926 & 0.932 & 0.150 & 0.129 & 0.119 \\
GET3D inv. w/ filt. + refine.   & \textbf{0.927} & 0.938 & \textbf{0.944} & \textbf{0.124} & \textbf{0.105} & \textbf{0.096}  \\
\bottomrule
\end{tabular}
}

\caption{\textbf{Ablation study results on ScanNet.} We ablate each component of \method{}: GAN backbone (EG3D vs GET3D), Phase I (filtered inversion) and Phase II (refinement). Using the GET3D backbone consistently outperforms an EG3D backbone. We also observe that both phase I and phase II generally boost performance, especially with more input views.}
\vspace{-2mm}
\label{table:ablation}
\end{table}

\begin{figure}[!tbp]
   \centering
      \includegraphics[width=\linewidth]{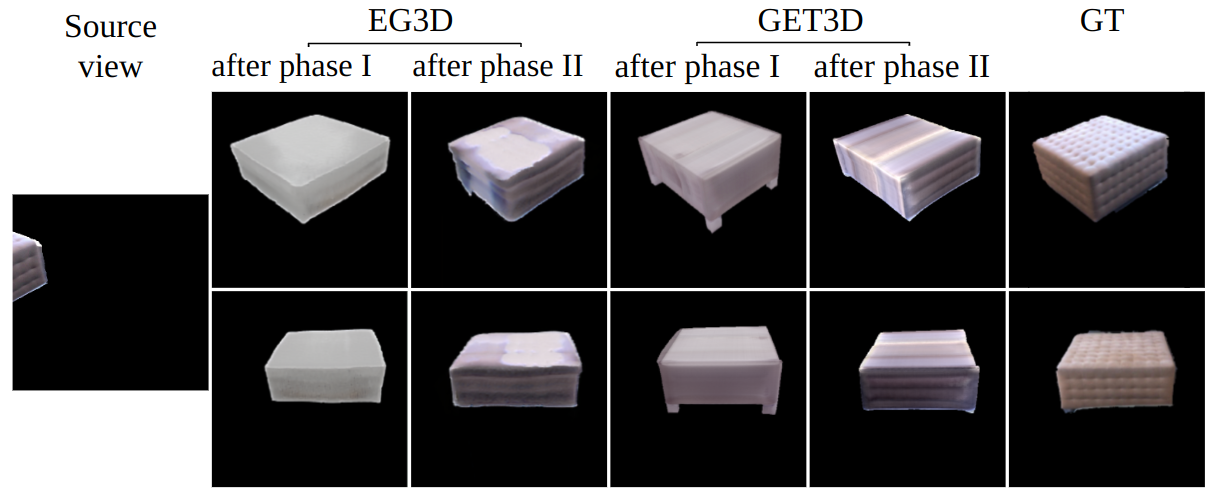}
     \caption{Visualization of \method results after Phase I (filtering inversion), and either before or after Phase II (refinement) for a given source view with EG3D and GET3D backbones.  Note that Phase II improves the texture and also removes extraneous legs.}
     \vspace{-2mm}
  \label{fig:ablation}
\end{figure}

\vspace{-2mm}
\paragraph{Comparison with Ablated \method.}
To answer \textbf{Q1}, we compare \method (using both GET3D and EG3D as the backbone) with and without the filtering process, and with and without the refinement phase. The results in Table \ref{table:ablation} show that the filtering process boosts overall performance. Qualitatively, we can see in Figure \ref{fig:ablation} that the filtered inversion phase fits the geometry and a rough texture from the source view. Then, in the refinement phase, the entire generator is fine-tuned to the input image to better fit the texture and also geometry.\looseness=-1

\vspace{-4mm}
\paragraph{GET3D versus EG3D.}
\method assumes GET3D as the backbone. To answer \textbf{Q2}, we adopt \method to use EG3D as the backbone.
The results in Table \ref{table:ablation} show that using GET3D as the backbone achieves the highest performance across all metrics. From Figure~\ref{fig:ablation}, we also observe that using GET3D yields a better reconstruction of the object during the filtered inversion phase (Phase I) compared to using EG3D, likely because the geometry and texture are disentangled in GET3D.
We also find that EG3D has a higher variance in performance across multiple runs when compared with GET3D. This suggests that conducting GAN inversion on GET3D is more stable than EG3D. Empirically, we observe that filtered inversion with the EG3D backbone has a higher chance of converging to a latent code that produces highly distorted reconstruction. 
Due to this instability in EG3D's inversion process, it benefits from the filtering process more than GET3D.\looseness=-1

\vspace{-3mm}
\paragraph{Rotational Delta Between Input and Target Views.}
To answer \textbf{Q3}, we plot LPIPS of various models against the rotational difference between the input and target views in Figure~\ref{fig:rot_error} (refer to section~\ref{sec:rot} in the supplementary material for plots of PSNR). We fit a line to the data points for each method. We find that Instant-NGP performs well on single-shot examples when the source and target view are extremely close to each other. However, as the target view gets further away from the source view, the performance of Instant-NGP's reconstruction degrades significantly as it does not have a prior on the geometry and texture of the object. We can also observe that FINV renders the target views consistently better than Instant-NGP~\cite{muller2022instant} and IBRNet~\cite{mueller2022autorf} when the source and target views are farther away.\looseness=-1

\begin{figure}[!tbp]
   \centering
      \includegraphics[width=\linewidth]{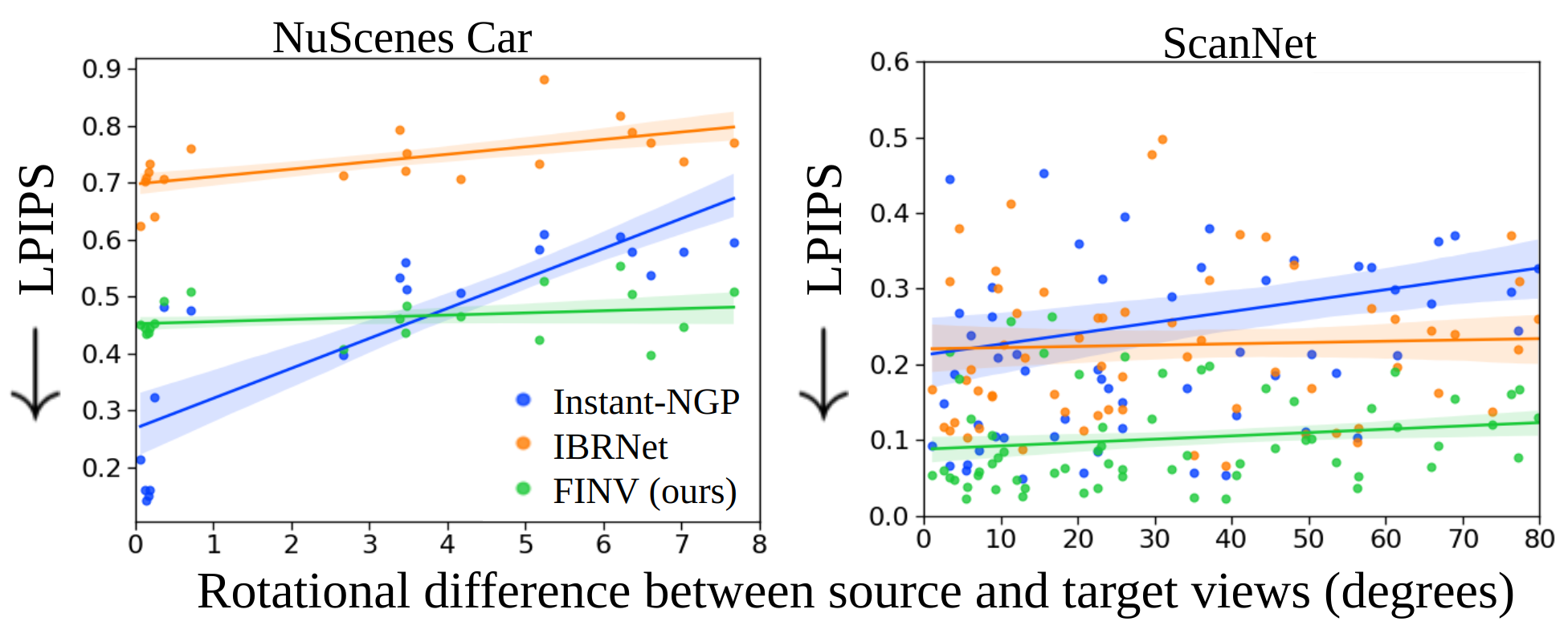}
     \vspace{-3mm}
     \caption{LPIPS as a function of the quaternion rotational difference between the source and target view. Each dot represents a data point, and the lines are linearly fitted to data points with linear regression. By learning object priors effectively, \method is less affected by the increase in rotational difference.}
     \vspace{-2mm}
  \label{fig:rot_error}
\end{figure}

\vspace{-2mm}
\section{Conclusion}
\vspace{-1.5mm}


In this paper we proposed Filtering Inversion (\method{}), 
a framework and optimization process that predicts a renderable 3D object representation from one or few \textit{partial} views. 
Through our experiments, we have shown that our method can be successfully applied to a variety of settings. 
We addressed shortcomings of previous works that are unable, when provided with partial views 
of the object of interest, to ``hallucinate'' unseen parts.
In contrast, our method produces complete novel views of the object. 
Our method generates a mesh without postprocessing, and it can process images sequentially, producing increasingly better results as more data becomes available.


\newpage
{\small
\bibliographystyle{ieee_fullname}
\bibliography{ref}
}

\clearpage
\newpage
\appendix
\section{Leveraging Synthetic Data}
Collecting 3D data in the real world is costly with current 3D scanning technology. Although our method is not limited to synthetic training data, we believe our ability to leverage priors from synthetic datasets to reconstruct real-world objects is an advantage. This allows us to use 3D model repositories from gaming, entertainment, and manufacturing industries.
It may also enable users of our method to generate more diverse data via data augmentation to cover the long-tail scenarios that are rarely encountered in the real world. From the table below, we can see that Objaverse~\cite{deitke2022objaverse}, a recently released dataset, consists of 818K synthetic 3D models over 21K object categories and is orders of magnitude larger than existing real-world 3D datasets such as Objectron~\cite{ahmadyan2021objectron} and CO3D~\cite{reizenstein2021common}. In this paper, we pre-trained our GANs on ShapeNet~\cite{chang2015shapenet}.

In light of the growing availability of 3D synthetic data, We believe that sim-2-real is an important path forward to leverage this data and push the frontier of real-world 3D perception, alongside parallel research that focuses on real data.

\begin{table}[h]
\centering
\resizebox{.8\linewidth}{!}{
\begin{tabular}{lllll}
\hline
               & Objectron & CO3D & Shapenet & Objaverse \\ \hline
\# objects    &    15K    &  19K    &   51K       &    818K       \\
\# classes &   9        &  50    &    55      &   21K        \\ \hline
\end{tabular}
}
\end{table}

\section{Datasets} \label{supp:dataset}
\noindent\textbf{Dataset Choice.} 
Contrary to many sparse-view reconstruction works that assume full 360-degree observational coverage of an object, our paper focuses on \textit{partial-view} circumstances (Figure 2) where 360-degree observations are typically not feasible due to real-world constraints, such as objects being placed against walls. We chose to evaluate our model on ScanNet instead of CO3D because, in CO3D, the videos are captured by placing the object on a solid surface and recording a full circle around it, ensuring that the entire object remains in view without any occlusion. However, these videos do not resemble natural observations like those found in ScanNet.
Since we are interested in reconstructing from real-world video observations, we prioritized evaluating our model on ScanNet instead of ShapeNet.

\noindent\textbf{Dataset Detail.}
ShapeNet~\cite{chang2015shapenet} contains 6778, 8443, and 7497 shapes for \textit{Chair}, \textit{Table}, and \textit{Car}, respectively. Following the experimental setting in GET3D~\cite{gao2022get3d}, we randomly choose 70\% of all shapes for training. For each shape, we render 24 images in Blender\footnote{We use the following rendering script: \url{https://github.com/nv-tlabs/GET3D/tree/master/render_shapenet_data}.}. Each image has a resolution of $128 \times 128$. 

For \textit{Chair} and \textit{Table}, we evaluate models on ScanNet~\cite{dai2017scannet}, a dataset of real-world indoor scene scans.
We select the scenes commonly used in other papers~\cite{yang2021objectnerf} and use all the chairs and tables observed in these scenes for evaluation. We do not filter sequences with inaccurate object masks or imperfect observations, as we want to evaluate the robustness of these models under noisy and partial measurements, which often occur when algorithms are deployed in the real world. When we evaluate the renderings, we only use images with accurate object masks. As ScanNet does not provide object poses, we use the poses estimated by Scan2CAD~\cite{avetisyan2019scan2cad}.
We adopt the same ScanNet scenes used in \cite{yang2021objectnerf}. Every ScanNet scene is a real-world scene that can contain many chair and table instances.

We evaluated all methods on all the chairs and tables that appear in 10 ScanNet scenes that are commonly chosen in related works such as Object-NeRF and NICE-SLAM. That adds up to 26 chairs and 18 tables. This is more than is often done, e.g., NodeSLAM evaluated 5 scenes and 10 objects for each of the three classes. 
Furthermore, we calculated the p-value statistical significance of our model's results against the results of one of the most competitive baselines (AutoRF) in Table 2.\footnote{Typically, p $<$ 0.05 indicates strong evidence that the difference is statistically significant.} 

\begin{table}[!h]
\centering
\resizebox{\linewidth}{!}{
\setlength{\tabcolsep}{5pt}
\begin{tabular}{@{}l c c c  c c c  c c c  }
\toprule
\multicolumn{1}{}{p-value} & \multicolumn{3}{c}{PSNR$\uparrow$} & \multicolumn{3}{c}{SSIM$\uparrow$} & \multicolumn{3}{c}{LPIPS$\downarrow$}  \\
 \textbf{\# views} & 1 & 3 & 5 & 1 & 3 & 5 & 1 & 3 & 5\\
\midrule
\textbf{ScanNet Chairs}  & 0.131 & 0.106 & 0.012 & $<$ 0.001 &$<$ 0.001 &$<$ 0.001 &$<$ 0.001 &$<$ 0.001 &$<$ 0.001 \\
\textbf{ScanNet Tables} & 0.001 & $<$ 0.001 & $<$ 0.001& $<$ 0.001& $<$ 0.001& $<$ 0.001& $<$ 0.001& $<$ 0.001 & $<$ 0.001 \\
\bottomrule
\end{tabular}
}
\end{table}

For \textit{Car}, we evaluate models on NuScenes~\cite{caesar2020nuscenes}, a driving dataset with 3D detection and tracking annotations. Following the experimental setting in \cite{mueller2022autorf}, we filter for sequences in the daytime and run a pre-trained 2D panoptic segmentation model~\cite{porzi2021improving} as NuScenes does not provide 2D segmentation masks. We also filter for sequences with sufficient camera movements and sufficiently clear observations.\footnote{We filter for observations above the resolution of $400 \times 400$ and sequences with a maximum rotational difference of more than 2 degrees.} Across all categories, we obtained around 50 sequences in total. We acknowledge the possibility of incorporating additional instances. However, due to time and computational limitations of some baselines, those in particular that require test-time optimization consume a significant amount of time.\looseness=-1

\section{Runtime Analysis} \label{supp:runtime}
We run our method on a Linux machine with NVIDIA A40 GPUs. In our experiments, we run 350 gradient update steps in the filtering inversion phase (Phase I), which takes approximately 21 seconds with a single A40 GPU and one input image of size $480 \times 480$. 
The reported 21-second runtime per frame corresponds to the results in the paper at 350 iterations, but it is not the minimum. In the table below we show the results with only 50 iterations or ~3 seconds. With a modest performance sacrifice, our method is suitable for near-real-time applications that process a new frame every few seconds, e.g., on a slow moving robot. Further acceleration is likely possible using optimization such as FP16 and fused operations.

\begin{table}[!h]
\centering
\resizebox{\linewidth}{!}{
\begin{tabular}{@{}l c c c  c c c  c c c  }
\toprule
& \multicolumn{9}{c}{\textbf{ScanNet Chairs}} \\
\multicolumn{1}{c}{} & \multicolumn{3}{c}{PSNR$\uparrow$} & \multicolumn{3}{c}{SSIM$\uparrow$} & \multicolumn{3}{c}{LPIPS$\downarrow$}  \\
 \textbf{\# views} & 1 & 3 & 5 & 1 & 3 & 5 & 1 & 3 & 5\\
\midrule
FINV-GET3D (3s) &  22.93 & 23.54 & 24.03 & 0.927 & 0.933 & 0.939 & 0.131 & 0.125 & 0.119 \\
FINV-GET3D (21s) & 24.61 & 24.96 & 26.23 & 0.937 & 0.944 & 0.950  & 0.102 & 0.089 & 0.082  \\
\bottomrule
\end{tabular}
}
\end{table}

\vspace{-2mm}
\section{More Visualizations on Rotational Delta}
\label{sec:rot}

\begin{figure}[!h]
   \centering
      \includegraphics[width=\linewidth]{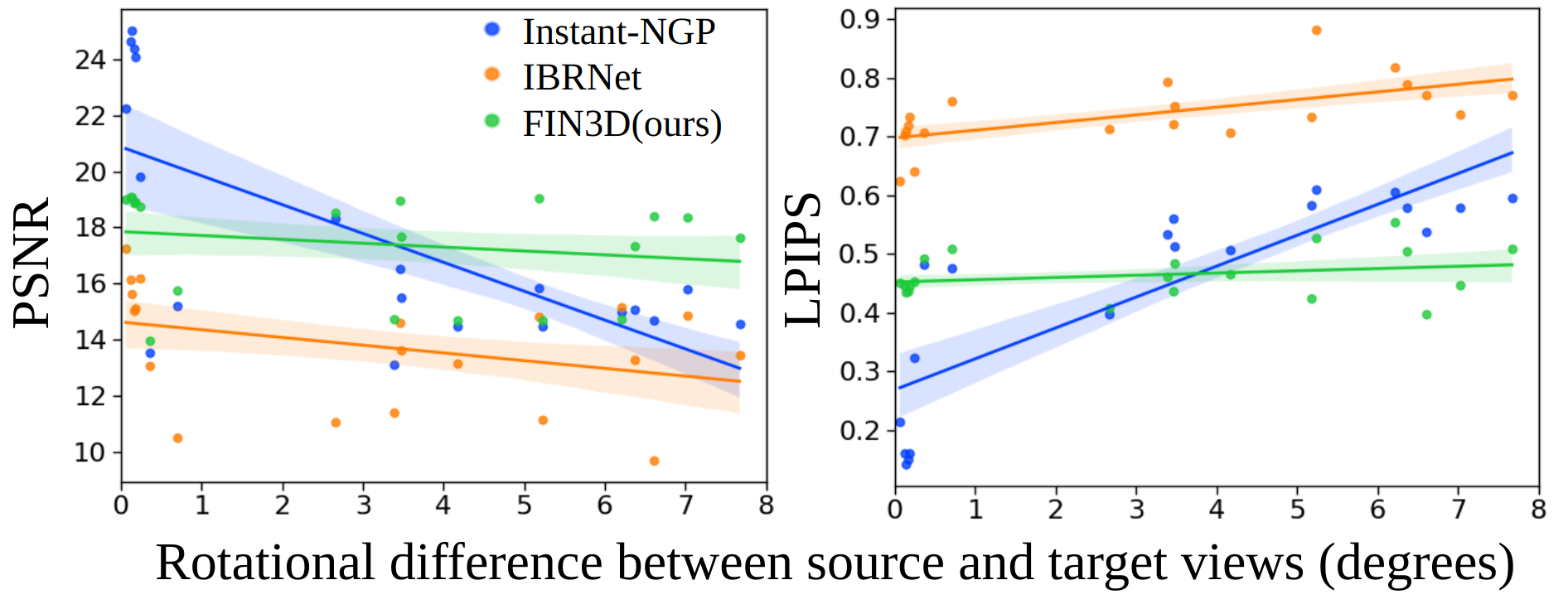}
     \caption{PSNR \& LPIPS as a function of the quaternion rotational difference between the source and target view. Each dot represents a data point in the NuScenes Car dataset, and the lines are linearly fitted to data points with linear regression.}
\end{figure}
\begin{figure}[!h]
   \centering
      \includegraphics[width=\linewidth]{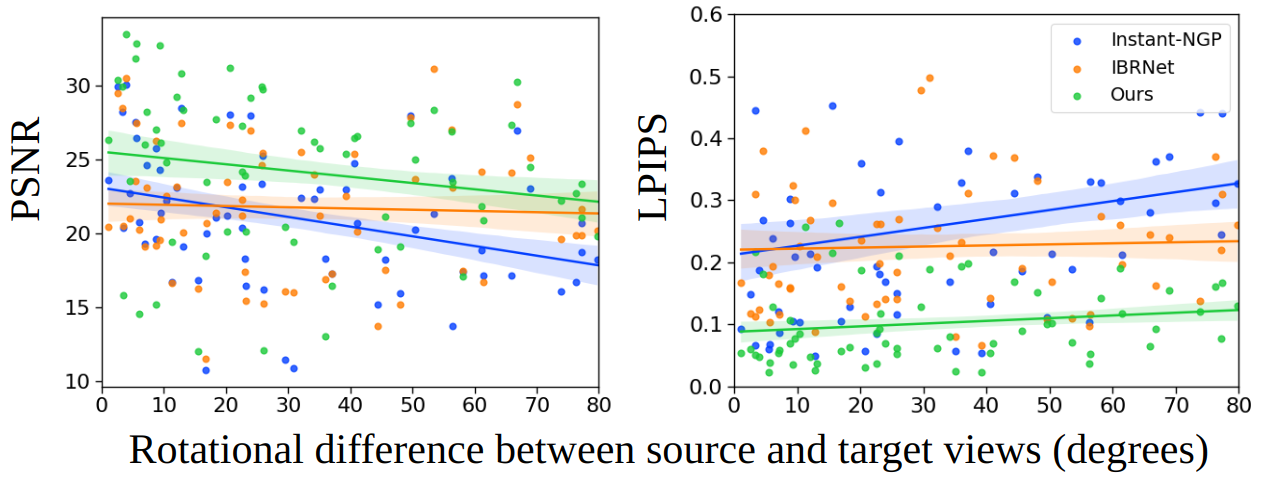}
     \caption{PSNR \& LPIPS as a function of the quaternion rotational difference between the source and target view. Each dot represents a data point in the ScanNet dataset, and the lines are linearly fitted to data points with linear regression.}
\end{figure}

\vspace{-2mm}
\section{Additional Analyses}
We provide additional reconstruction results of our model in Figure~\ref{fig:additional}. We wondered how our system, having seen only synthetic objects, would handle uncommon objects. Combining learned object priors and the use of the Refinement phase in \method, we find that it can reconstruct, from a single view, the avocado chair geometrically better than methods that do not learn object priors such as NeRS~\cite{zhang2021ners} and Differentiable Stereopsis (DS)~\cite{goel2022differentiable}. 
\begin{figure}[!h]
   \centering
      \includegraphics[width=\linewidth]{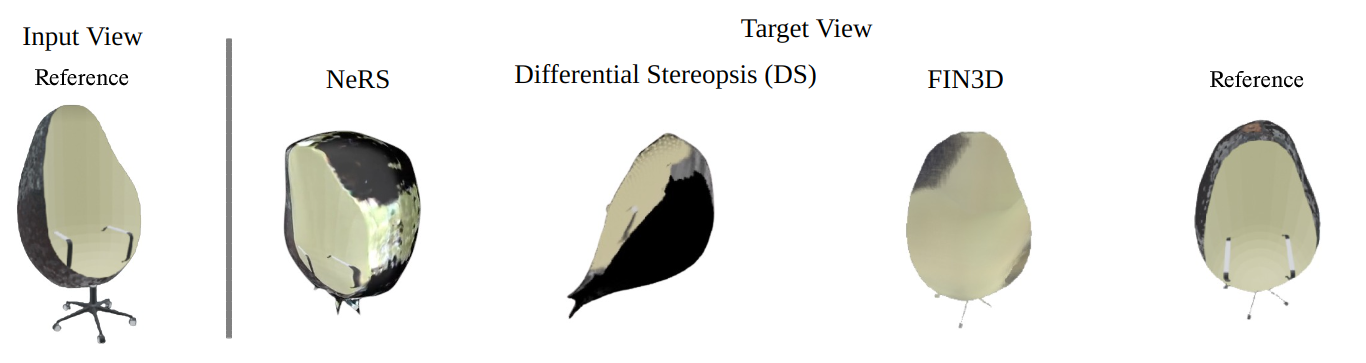}
     \caption{Visualization of our method (\method) and baselines on an avocado chair, viewed from a different angle, after training on a single image.} 
\end{figure}
\begin{figure}[!h]
   \centering
    \includegraphics[width=\linewidth]{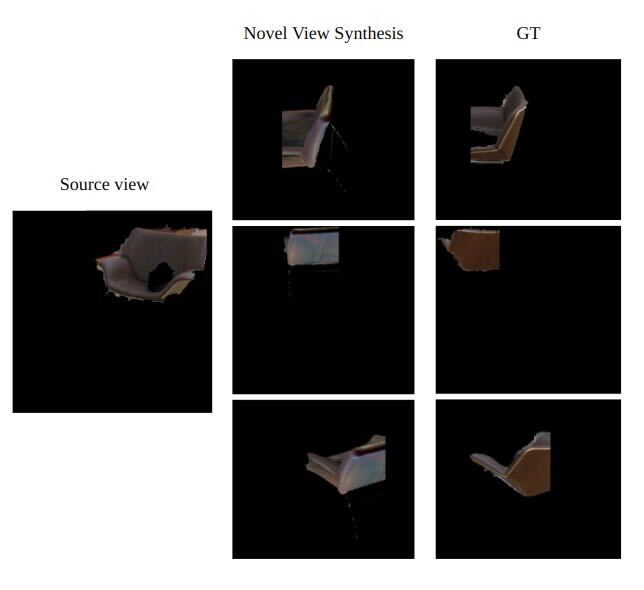}
         \caption{One-shot reconstruction results of \method. In this example, we can observe that our model produces decent textures for the observed sides of the chair but unrealistic textures for the unobserved side of the chair.}
  \label{fig:texture}
\end{figure}

While \method makes a step towards partial-view object-centric reconstruction, it has some limitations. Chairs generated by a trained GET3D model tend to have consistent textures. That is, the model has the prior that, for instance, the back side of a chair tends to have the same texture as the front side of the chair. However, this prior can be lost or altered when we fine-tune the entire generator to fit the input observations (during the Refinement phase), resulting in unrealistic textures for the unobserved sides of the object. Figure~\ref{fig:texture} gives one such example.

Additionally, \method does not model any optical phenomena such as specular highlights, reflections, and transparency. When generating textured objects, the texture then has baked-in light, one important step forward for generative objects from few views would be to include the physical proprieties of the object texture, \textit{e.g.}, roughness. This is still an open problem in the graphics community whereas differentiable renderers are used to optimize shape, texture, material, and environment light map~\cite{Hasselgrenrender}. A promising extension is to use DIB-R++~\cite{chen2021dib} to predict environmental map light and combine that with GET3D~\cite{gao2022get3d} to generate view-dependent lighting effects.

\begin{figure}[!h]
   \centering
    \includegraphics[width=\linewidth]{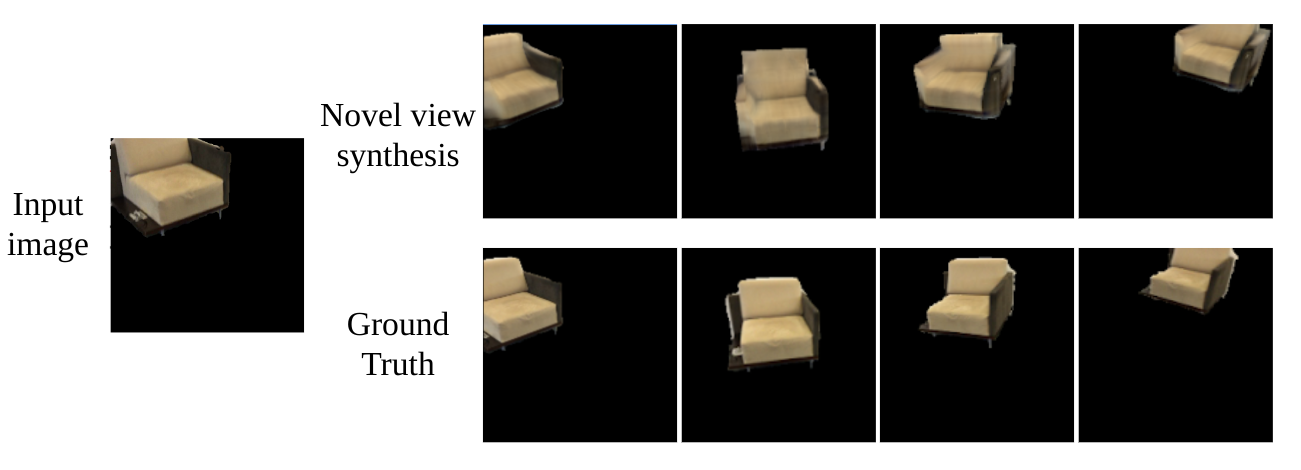}
     \includegraphics[width=\linewidth]{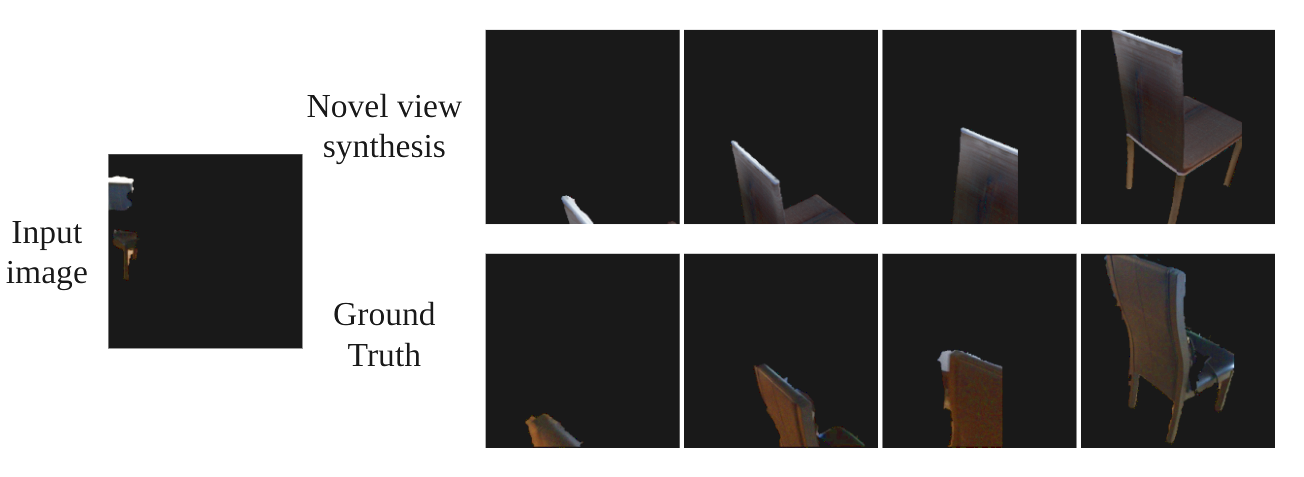}
      \includegraphics[width=\linewidth]{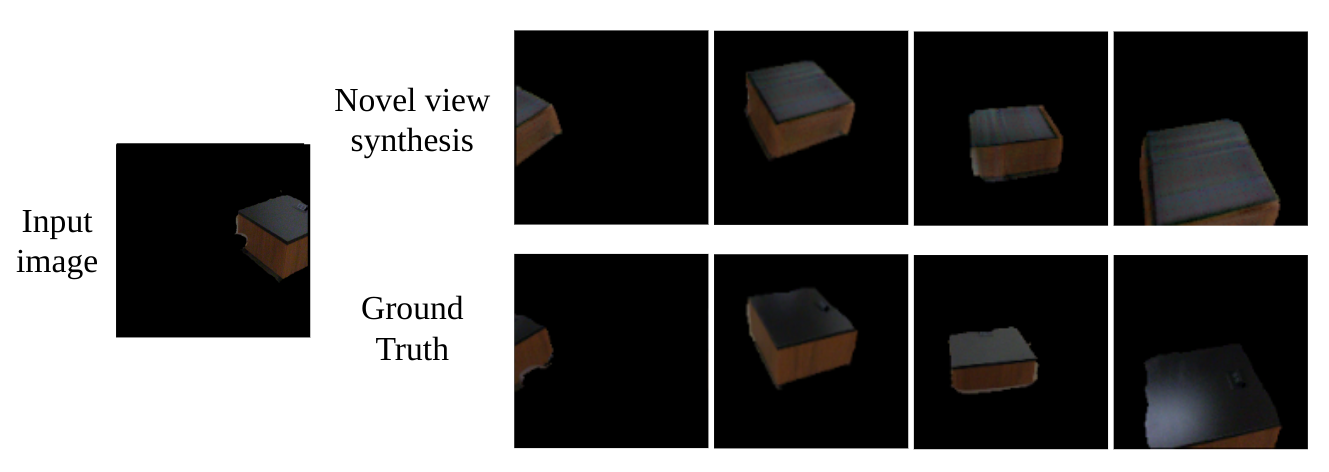}
       \includegraphics[width=\linewidth]{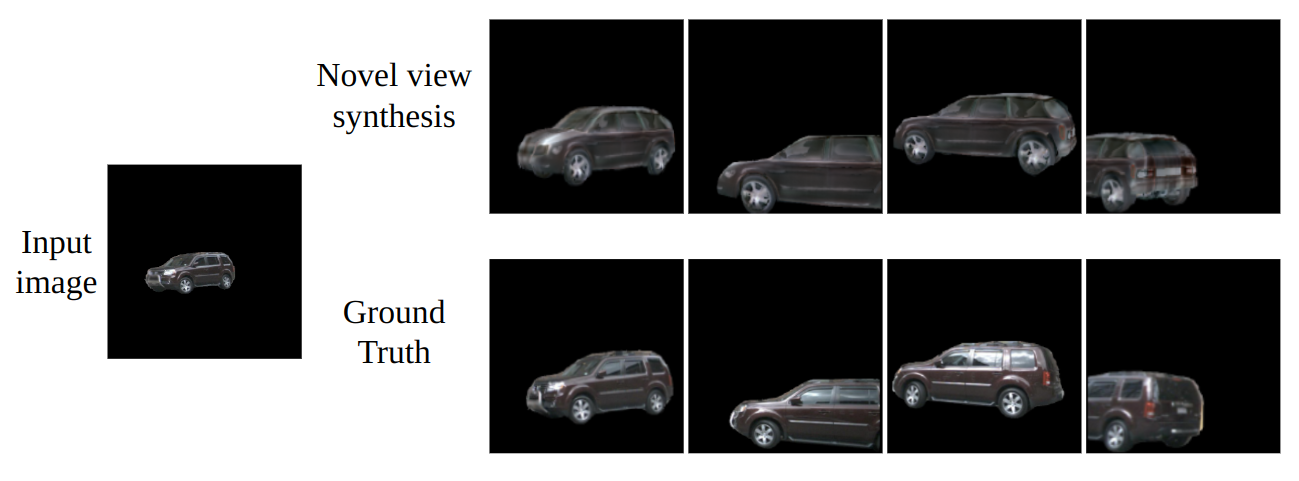}
              \includegraphics[width=\linewidth]{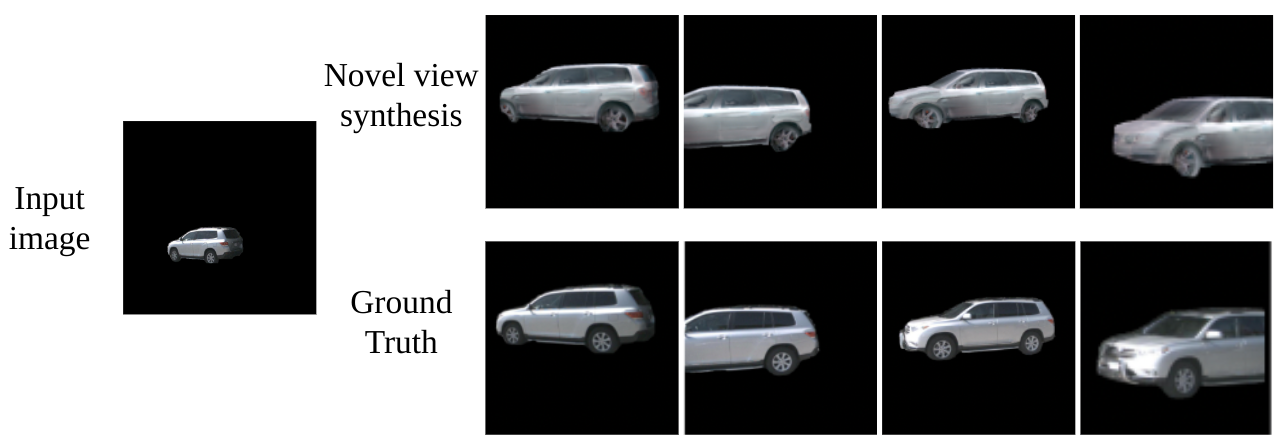}

         \caption{Additional reconstruction results of \method.}
  \label{fig:additional}
\end{figure}
\vspace{-2mm}


\vspace{-2mm}
\section{More Related Works} \label{supp:related}
 %
Novel view synthesis, a long-standing research problem in the field of computer vision, entails constructing new views of a scene or an object from one or more views.
%
Recent works demonstrate the effectiveness of learned implicit neural representations for rendering novel views ~\cite{lombardi2019neural,mildenhall2020nerf,muller2022instant,sitzmann2019scene,yang2021learning}. However, these approaches require many input views and substantial optimization time per scene as they fit a single model to each scene or object.
Recent work has explored 3D shape and/or texture generation~\cite{zhang2021ners,yariv2021volume,huang2020cvpr}, these methods assume views that fully cover the object, whereas our method works with a single input view. 
%
Additionally, locally-conditioned CNN features have been used to generalize neural implicit representations across scenes~\cite{saito2019pifu,peng2020convolutional,trevithick2020grf,yu2021pixelnerf,wang2021ibrnet,chen2021mvsnerf}.
These methods have shown excellent performance in novel-view synthesis in a range of testing scenarios. However, they mostly learn pixel-level priors and do not consider generative object-level priors. This limits their generalization capability, especially in cases with \textit{partial} observations.

There is abundant work on learning to map image to 3D representations such as voxels, point cloud, and mesh. For example,
\cite{choy20163d} takes the image as input and generates a 3D object by voxel. \cite{fan2017point} maps an image to a point cloud by learning a residual mapping in the latent space of an autoencoder. \cite{wei2022flow} uses a GAN architecture to generate 3D point cloud from a 2D image. In recent years, image to mesh methods are growing in popularity. \cite{wang2018pixel2mesh} learns a mapping from the image to mesh by using a graph-based convolutional neural network.
Orthogonal to the 3D representation is to incorporate multiple views. 3D-R2N2~\cite{choy20163d} proposes to update its reconstruction given new views using an LSTM. In this work, we instead focus on implicit reconstruction, where the goal is to synthesize new views of a captured object.

Another related work to our paper is NodeSLAM~\cite{sucar2020nodeslam}, which considers generative object priors (i.e., class-conditional variational autoencoder) for object shape reconstruction based on RGBD inputs. However, NodeSLAM requires depth observation and only reconstructs geometry.
A more recent method, AutoRF~\cite{mueller2022autorf}, also uses a pre-trained 3D generative prior to reconstruct objects in the wild (specifically cars in street scenes) from single RGB observation. However, AutoRF uses a separate ResNet encoder network to map the input images to a latent code, which may not generalize well to out-of-distribution data not seen during training. That is, it would be hard for AutoRF to leverage the large repository of synthetic data to reconstruct real-world objects.
In contrast, FIN3D performs filtered inversion through a pre-trained 3D generator and utilizes a Refinement Phase to address the simulation-to-real gap.

\section{Future Work} \label{supp:future}
We have made progress in generating photo-realistic views of objects that were not fully visible, 
 important challenges remain. 
One important aspect of our proposed system is based on leveraging a generative method (GET3D). Generative methods can be quite challenging and time-consuming to train. We expect research in these areas to improve the accessibility of our proposed method. 
Also, our method currently focuses on single categories, exploring larger object diversity will potentially broaden the applicability of our method. 
This is somewhat of an open problem as most 3D GAN methods target single categories.
The increasing availability of large datasets of 3D models~\cite{deitke2022objaverse}, as well similar real-world data, will facilitate this type of research.\looseness=-1



\end{document}